\documentclass[journal,comsoc]{IEEEtran}
\usepackage[T1]{fontenc}% optional T1 font encoding

\usepackage{pifont}
\usepackage{multirow}
\usepackage{color}
\usepackage{setspace}
\usepackage{cite}
\ifCLASSINFOpdf
  \usepackage[pdftex]{graphicx}
\else
\fi
\usepackage{amsmath}
\usepackage[cmintegrals]{newtxmath}
\usepackage{algorithmic}
\usepackage{array}
\usepackage{url}
\begin{document}
\title{Contrastive Semi-supervised Learning for Domain Adaptive Segmentation Across Similar Anatomical Structures}

\author{Ran~Gu, Jingyang~Zhang, Guotai~Wang, Wenhui~Lei, Tao~Song, Xiaofan~Zhang, Kang~Li and Shaoting~Zhang
\thanks{This work was supported by National Natural Science Foundation of China (No.61901084 and No.81771921), and by Department of Science and Technology of Sichuan Province, China (No.20ZDYF2817). The work was done during R. Gu's internship at SenseTime Research. R. Gu and J. Zhang contributed equally to this work. (corresponding author: guotai.wang@uestc.edu.cn).}
% \thanks{The next few paragraphs should contain 
% the authors' current affiliations, including current address and e-mail. For 
% example, F. A. Author is with the National Institute of Standards and 
% Technology, Boulder, CO 80305 USA (e-mail: author@boulder.nist.gov). }
\thanks{R. Gu, G. Wang and S. Zhang are with the School of Mechanical and Electrical Engineering, University of Electronic Science and Technology of China, Chengdu, China, and also with the Shanghai AI Lab, Shanghai, China. S. Zhang is also with the SenseTime Research, Shanghai, China.}
\thanks{J. Zhang is with the School of Biomedical Engineering, Shanghai Jiao Tong University, Shanghai, China, and also with the School of Biomedical Engineering, ShanghaiTech University, Shanghai, China.}
\thanks{W. Lei and X. Zhang are with the School of Electronic Information and Electrical Engineering, Shanghai Jiao Tong University, Shanghai, China, and also with the Shanghai AI Lab, Shanghai, China.}
\thanks{T. Song is with the SenseTime Research, Shanghai, China.}
\thanks{K. Li is with the West China Hospital, Sichuan University, Chengdu, China.}
}

\markboth{}
{Gu \MakeLowercase{\textit{et al.}}: Contrastive Semi-supervised Learning for Domain Adaptive Segmentation Across similar Anatomical Structures}



\maketitle

\maketitle
\begin{abstract}
%Domain shift problem is attracting increasing interests in medical image analysis since there exists performance gap between source and target domain acquired with different modalities or protocols. Most of the recent works are dedicated to diminishing these domain gaps on data modalities, phases, or scan-sequences among the same field-of-view. The difference between two domains is mainly reflected in the distribution of pixel intensity, and previous Unsupervised Domain Adaptation (UDA) methods have been achieving great progress by introducing constraint functions or adversarial training. In this work, we built domain shift in a totally different perspective, in which two datasets were acquired on total different organs while %only have share similar anatomical structures. We believe that transferring knowledge from widely-available public labeled datasets to an unique datasets is a promising way to improve the annotation efficiency due to their similar anatomical structures. 
Convolutional Neural Networks (CNNs) have achieved state-of-the-art performance for medical image segmentation, yet need plenty of manual annotations for training. Semi-Supervised Learning (SSL) methods are promising to reduce the requirement of annotations, but their performance is still limited when the dataset size and the number of annotated images are small. Leveraging existing annotated datasets with similar anatomical structures to assist training has a potential for improving the model's performance. However, it is further challenged by the cross-anatomy domain shift due to the different appearance and even imaging modalities from the target structure. To solve this problem, we propose Contrastive Semi-supervised learning for Cross Anatomy Domain Adaptation (CS-CADA) that adapts a model to segment similar structures in a target domain, which requires only limited annotations in the target domain by leveraging a set of existing annotated images of similar structures in a source domain. We use Domain-Specific Batch Normalization (DSBN) to individually normalize feature maps for the two anatomical domains, and propose a cross-domain contrastive learning strategy to encourage extracting domain invariant features. They are integrated into a Self-Ensembling Mean-Teacher (SE-MT) framework to exploit unlabeled target domain images with a prediction consistency constraint. Extensive experiments show that our CS-CADA is able to solve the challenging cross-anatomy domain shift problem, achieving accurate segmentation of coronary arteries in X-ray images with the help of retinal vessel images and cardiac MR images with the help of fundus images, respectively, given only a small number of annotations in the target domain. %Our code will be available at~\url{https://github.com/HiLab-git/DAG4MIA}
\end{abstract}


\section{Introduction}
\label{sec1:introduction}
% for semi-supervised learning (mean teacher) + BN
\IEEEPARstart{R}{ecently}, Convolutional Neural Networks (CNNs) have achieved remarkable progress in medical image segmentation\cite{shen2017deep,gu2020canet}, yet requiring a large amount of manual annotations for training images, which is highly time-consuming and labor-intensive to collect. Therefore, it is desired to reduce the manual annotations for model training while maintaining the segmentation performance. Semi-Supervised Learning (SSL) has been widely used to reduce the 
required annotations, as it only requires a small set of labeled data with the availability of a large set of unlabeled data~\cite{tajbakhsh2020embracing}. 
%Recent commonly adopted SSL methods include entropy minimization~\cite{grandvalet2005semi} that encourage model to make low entropy prediction on unlabeled data and abundant consistency regularization methods~\cite{sajjadi2016regularization} that encourage model make consistent prediction between transformed data and original data. Specifically, the Mean-Teacher (MT) model working with large unlabeled datasets and online guiding training makes great success through enforcing the consistency of predictions from perturbed inputs between student and teacher models~\cite{tarvainen2017mean}. 
Although SSL methods have achieved promising performance in medical image segmentation~\cite{sajjadi2016regularization,tarvainen2017mean}, they often rely on a very large set of unannotated images and the performance is still limited when the number of labeled images is very small. For a lot of medical imaging applications, it is not only time-consuming to acquire annotations, but also difficult and expensive to collect a large set of unannotated images, leading to a small set of available training samples~\cite{tajbakhsh2020embracing}. In such cases, it is challenging for most existing SSL methods to achieve high performance with a small training set of which part samples are annotated. 

Although the small set of training samples as well as lack of annotations for a given task are common problems in the field of medical image segmentation, there are many existing datasets with full annotations. It is desirable to leverage such datasets to assist the training of a model for a target task, i.e., adapting a model trained with existing datasets (i.e., source domain) to a specific target dataset (i.e., target domain). As the source and target domain images are often acquired with different imaging protocols, such as different modalities, contrasts, patient groups and even anatomical structures, the performance is limited if such pre-trained models are directly applied to the target domain images for inference. Recently, Domain Adaptation (DA) is attracted increasing attentions, which assumes that the same task is involved in the source and target domains, e.g., adapting a model trained with annotated heart Magnetic Resonance Images (MRI) to segment heart Computed Tomography (CT) images~\cite{chen2019synergistic}. To alleviate the performance gap between the two domains, a widely used method is to fine-tune the pre-trained models with target domain images~\cite{Tajbakhsh2016fine}. However, the fine-tuning process requires annotations in the target domain and cannot leverage unannotated images for training. Alternatively, Unsupervised Domain Adaptation (UDA) methods have been increasingly investigated to adapt a model trained with a source domain to a target domain. UDA methods do not require annotations in the target domain and usually translate  source-domain images~\cite{chen2019synergistic} to target domain-like images or learn domain-invariant features~\cite{dou2018unsupervised} to achieve good results.
% for domain adaptation
%Domain shift problem is attracting increasing interests in medical image analysis since there exists performance gap between source and target domain acquired with different modality or protocols~\cite{guan2021domain}. Most of the recent works are dedicated to diminishing these performance gaps and proposed Unsupervised Domain Adaptation (UDA) methods~\cite{ganin2015unsupervised,wang2018deep}. The main purpose of unsupervised domain adaptation (UDA) studies how to transfer knowledge from a labeled source domain to an unlabeled target domain. Existing UDA methods can be mainly summarised as two aspects that one parts of approaches aim to minimize the distance between source and target domain on a latent feature space~\cite{chen2019joint,Zhu2019adapting}, and another mainstream approaches are based on Generative Adversarial Networks (GANs)~\cite{tzeng2017adversarial,zhu2017unpaired}. Recently, most of the state-of-the-art UDA methods are focusing domain discrepancy either on the cross-style in natural image or cross-modality in medical imaging~\cite{zhu2017unpaired,chen2019synergistic}.

\begin{figure}[t]
    % \vspace{-0.2cm}
    % \setlength{\abovecaptionskip}{-0.2cm}
    % \setlength{\belowcaptionskip}{-0.2cm}
    \centerline{\includegraphics[width=\columnwidth]{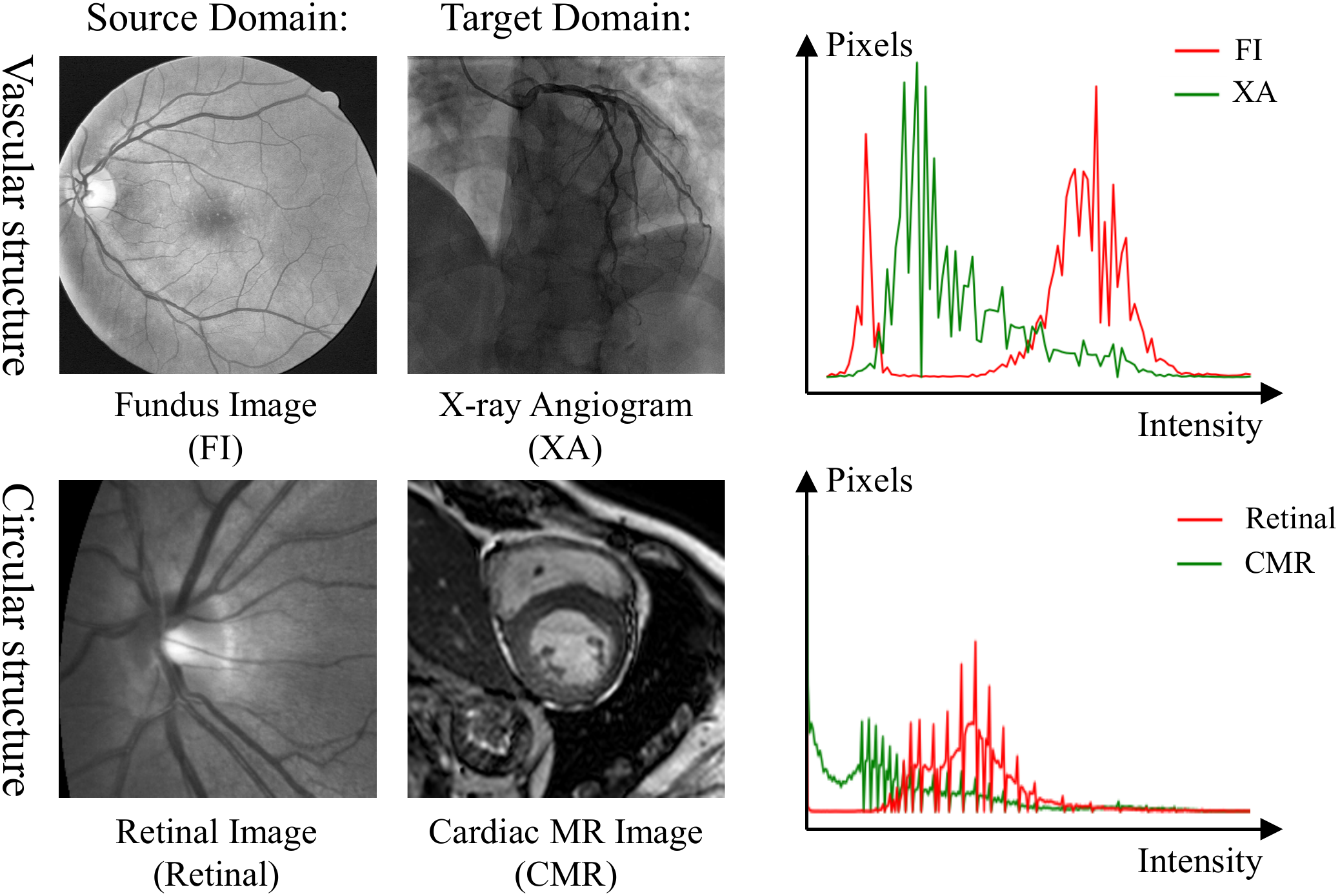}}
    \caption{Illustration of the challenging cross-anatomy domain shift in two scenarios.
    Row 1 shows the domain shift between fundus image (FI) and X-ray angiogram (XA) with similar vascular structures. Row 2 shows the cross-circular domain shift between Retinal image (Retinal) and Cardiac MR image (CMR) with circular structures. Note the different intensity histograms, morphologies, and contexts of the similar anatomical structures in the source and target domains.}
    \label{fig1:domain_shift}
\end{figure}

However, most of the state-of-the-art DA methods for medical image segmentation require that the source and target domains have the same set of anatomical structures even they can be from different imaging modalities~\cite{chen2019synergistic}. Such a requirement prevents these methods from leveraging images of other anatomical structures for training, and unlocking this requirement would enlarge the scope of candidate source domains, which helps to improve the segmentation in the target domain when a source domain with exactly the same anatomical structures is not available.
%Few works focus on exploring the connection of the inner components between source and target domains, especially across anatomy in medical imaging. In common perception, owing to the same anatomical structures in medical images can easily be recognized by humans even they are not professional radiologists. 
For instance, for segmentation of coronary arteries from 2D X-ray Angiogram (XA) with limited annotations, it is hard to find an annotated dataset with the same structure as the target domain. However, there are a lot of public fundus images with annotated retinal vessels (e.g., DRIVE~\cite{DRIVE} and STARE~\cite{STARE}), where the retinal vessels share similar tubular structures with the coronary arteries, as shown in fist row of Fig.~\ref{fig1:domain_shift}. Another example is the similar circular structures between Left Ventricle blood cavity (LV) and the Myocardium (Myo) in Cardiac MR (CMR) images and the optic cup and disc in public Retinal images (Retinal), as shown in the second row of Fig.\ref{fig1:domain_shift}. Hence, it is promising to transfer the knowledge from these retinal vessel datasets to the coronary artery segmentation task~\cite{yu2019annotation}, as they can be regarded as cost-free source domain images. However, the different morphologies and contexts of these two kinds of vessels make it hard to achieve accurate results for existing UDA methods that are designed to deal with the same anatomical structures.

Unlike previous works, we investigate the cross-anatomy semi-supervised domain adaptation problem for medical image segmentation. Compared with existing domain adaptation methods that require the segmented objects to be the same in the source and target domains, our method relaxes this requirement and enables adapting a model to segment a different anatomical structure to the source domain. Compared with existing semi-supervised methods requiring the training samples from the same domain, our method leverages images from other domains with similar anatomical structures to assist the training process, which can improve the segmentation performance in the target domain. %To effectively make use of the knowledge from the source domain, we utilize a simple but effective method, in which we share the convolutional layers across domain aiming at collaboratively capturing comprehensive domain-invariant shape information because of similar anatomical structure. Meanwhile, we introduce specific Batch Normalization (BN) across anatomy domain for maintaining high discriminative domain-specific characteristics. In addition, we propose a contrastive learning strategy to improve the model's feature representation ability considering the small set of target domain images. To perform cross-anatomy contrastive learning and match the features of the similar anatomical structure, we define the two domain features fed into their corresponding domain-specific batch normalization as the positive pairs, while any feature fed into their anisotropic batch normalization can form one sample of negative pairs.
% make summary

Our proposed method is referred to as Contrastive Semi-supervised learning for Cross Anatomy Domain Adaptation (CS-CADA) for medical image segmentation. The main contributions are:

1) To reduce the annotation cost and overcome the problem of limited training images in a target domain, we propose to leverage an existing annotated dataset of a similar anatomical structure from a different domain for training. A novel framework called CS-CADA that is a generalization of existing semi-supervised and domain adaptation methods is introduced. 

2) To deal with domain shift between the two domains while transferring the knowledge of similar anatomical structures, we use Domain-Specific Batch Normalization (DSBN)~\cite{ganin2016domain} with shared convolutional kernels for the two domains, and integrate it into a Mean Teacher (MT)-based consistency regularization framework to leverage unannotated images in the target domain. %Based on above these, we previously proposed Semi-Supervised Cross Anatomy Domain Adaptation (SS-CADA) for cross vessel structure segmentation~\cite{zhang2021sscada}.

3) To better learn domain-invariant features, we propose a cross-domain contrastive learning strategy, where a pair of  features from the two domains based on their individual DSBN are treated as positive pairs and otherwise negative pairs.

4) Extensive experiments on two different scenarios (coronary artery and cardiac MRI segmentation with the assistance of retinal vessel images and fundus images, respectively) demonstrate that our proposed CS-CADA achieved accurate segmentation results with better performance than existing semi-supervised and domain adaptation methods when annotations in the target domain are limited. 

A preliminary version of this work was published in 2021~\cite{zhang2021sscada}, where we applied the combination of DSBN and SE-MT for coronary artery segmentation. In this extension, we provide detailed descriptions of our framework, and introduce the cross-domain contrastive learning strategy for better performance. The method has also been validated with more extensive experiments of different applications in this work.

\section{Related works}
\label{sec2:related_works}
\subsection{Semi-supervised Learning}
Semi-supervised learning methods have been widely used to reduce the amount of required annotations for medical image segmentation~\cite{tajbakhsh2020embracing}. Existing methods mainly use strategies such as pseudo label~\cite{chen2021semi}, adversarial training~\cite{zhang2017deep} and consistency regularization~\cite{sajjadi2016regularization}. Pseudo label-based methods train a model on annotated images to generate masks  for unannotated images that are then used to update the segmentation model. To improve the quality of pseudo labels, different strategies including randomly selected propagation~\cite{fan2020inf} and uncertainty-based refinement~\cite{wang2021semi} have been proposed. Adversarial learning-based methods~\cite{zhang2017deep} use discriminators to encourage the predictions of unannotated images to be similar to those of the annotated ones, where prior information~\cite{zheng2019semi} may be used to improve the performance. Consistency regularization methods encourage  predictions from one input image under different perturbations to be consistent, e.g.,  transformation consistency~\cite{bortsova2019semi}, dual-task consistency~\cite{luo2021semi}, and perturbation-based consistency~\cite{yu2019uncertainty}.

Mean teacher is a widely used consistency-based SSL method~\cite{tarvainen2017mean}. It uses a self-ensembling of a student model as the teacher model, and encourages consistent predictions between the two models~\cite{cui2019semi}. It has been extended to Uncertainty Aware Mean-Teacher (UA-MT)~\cite{yu2019uncertainty} and combined with transformation consistency to achieve better results~\cite{li2020transformation} for medical image segmentation. %With the mean teacher proposed for semi-supervised learning~\cite{tarvainen2017mean}, a lot of mean teacher based methods are proposed for medical image segmentation. Cui et al.~\cite{cui2019semi} adapted mean teacher model for brain lesion segmentation by introducing a segmentation consistency loss to constraint model producing consistent output for the same image under different resolutions. Yu et al.~\cite{yu2019uncertainty} further extended mean teacher model with uncertainty guided map to encourage model focusing on hardly determining areas for left atrium segmentation. Li et al.~\cite{li2020transformation} introduced tansformation-consistent strategy including rotation, flipping, and scaling operations in the self-ensembling model to enhance the regularization effect for medical image segmentation. %Mean teacher based semi-supervised method is also adapted to tackle cross-modality domain shift problem. 
However, most existing SSL methods assume the training samples are from the same domain and hardly obtain high performance when the available images and annotations are limited. In this work, we leverage annotated images from an existing dataset with similar structures (e.g., a public dataset) to assist the semi-supervised training for better performance.
\subsection{Transfer Learning and Domain Adaptation}
Transfer Learning (TL) aims to transfer knowledge learned from a source dataset to deal with data in a new target dataset~\cite{pan2009survey}. Early transfer learning methods mainly fine-tune part or the whole set of parameters in a pre-trained model with an annotated target dataset~\cite{yosinski2014transferable,tajbakhsh2016convolutional}. Chen et al.~\cite{chen2015standard} fine-tuned a pre-trained CNN for localizing standard planes in ultrasound images. Tajbakhsh et al.~\cite{tajbakhsh2016convolutional} demonstrated that knowledge can be transferred from natural images to medical images through fine-tuning. Although fine-tuning is easy to implement straightforward for transfer learning, it is always faced with three main issues, i.e., where, how and when to fine-tune~\cite{pan2009survey}. If the source and target datasets have a large gap, the fine-tuning is less effective. Differently from classical transfer learning, Domain Adaptation (DA) deals with the same set of target objects in different domains (e.g., imaging protocols, patient groups, or intensity distribution), which is more effective to transfer the knowledge in the source domain to the target domain due to the shared anatomical structure. In addition, it can leverage the source and target images simultaneously for training, rather than using the source and target images in two independent stages in typical transfer learning.

Domain Adaptation has been widely used to alleviate the performance degradation when the distribution of target data differs from that of source data~\cite{ganin2015unsupervised,tzeng2017adversarial}. DA methods mainly have three categorises based on the type of annotations. The first is fully-supervised DA, where fine-tuning is the most representative technique for adapting a trained model to the target domain with full annotations~\cite{Yosinski2014how,valverde2019one}. The second is weakly or semi-supervised DA where only coarse or partial annotations in the target domain are used for model adaptation~\cite{chen2021adaptive}. For example, Dorent et al.~\cite{dorent2020scribble} employed scribbles in the target domain to perform model adaptation for vestibular schwannoma segmentation. Li et al.~\cite{li2020dual} used a dual-teacher semi-supervised domain adaptation method when a small ratio of the target domain images have annotations. %Chen et al.~\cite{chen2021adaptive} introduced adversarial learning for semi-supervised left atrium segmentation on cross modality data. 
Additionally, Unsupervised Domain Adaptation (UDA) does not require annotations in the target domain. 
UDA methods usually use Generative Adversarial Networks (GAN) to achieve image-level or feature-level alignment between the source and target domains. For example, Zhu et al.~\cite{zhu2017unpaired} used Cycle-GAN to convert source domain images to target-domain like images to reduce the domain gap. Kamnitsas et al.~\cite{kamnitsas2017unsupervised} and Dou et al.~\cite{dou2018unsupervised} used GAN to obtain domain invariant features for adaptation. The Simultaneous Image and Feature Alignment (SIFA)~\cite{chen2019synergistic} combined the advantage of these two categories and has achieved state-of-the-art performance for UDA task. 

Although existing annotation-efficient DA methods achieved promising performance in the target domain with limited annotations, most of them can only deal with two domains with the same anatomical structure, and are not applicable for cross-anatomy domain adaptation.
In this work, we deal with the domain shift between two datasets of different anatomical structures with similar shapes.
\subsection{Contrastive Learning}
Contrastive learning commonly employs a contrastive loss to enforce representations to be similar for positive pairs and dissimilar for negative pairs~\cite{hadsell2006dimensionality}. Previous contrastive learning methods are mainly proposed as a self-supervised pre-training strategy to train a powerful and representational feature extractor that can be fine-tuned for down-stream tasks~\cite{chen2020simple,chaitanya2020contrastive}. %Chen et al.~\cite{chen2020simple} proposed a simple contrastive learning framework and evaluated the effectiveness of utilizing stronger data augmentation when constructs the positive and negative pairs. Chaitanya et al.~\cite{chaitanya2020contrastive} further used contrastive learning of global and local features sequentially for 3D volume medical image segmentation with limited annotations. Almost of works based on contrastive learning are dedicated to conducting self-supervised training under limited labels~\cite{chen2020simple,lei2020one}.

Recently, contrastive learning has also been used for domain adaptation. Kang et al.~\cite{kang2019contrastive} proposed an end-to-end contrasitve adaptation network that minimizes the intra-class domain discrepancy and maximizes the inter-class discrepancy. Singh et al.~\cite{singh2021clda} employed class-wise and instance-level contrastive learning to respectively minimize the inter-domain and inter-domain discrepancy in semi-supervised domain adaptation. In this work, we design a novel contrastive learning strategy for cross-anatomy domain adaptation, which encourages the model to extract comprehensive domain-invariant features across similar anatomical structures.
\begin{figure*}[t]
    \centering
    % \vspace{-0.2cm}
    % \setlength{\abovecaptionskip}{-0cm}
    % \setlength{\belowcaptionskip}{-0.2cm}
    \includegraphics[width=0.85\textwidth]{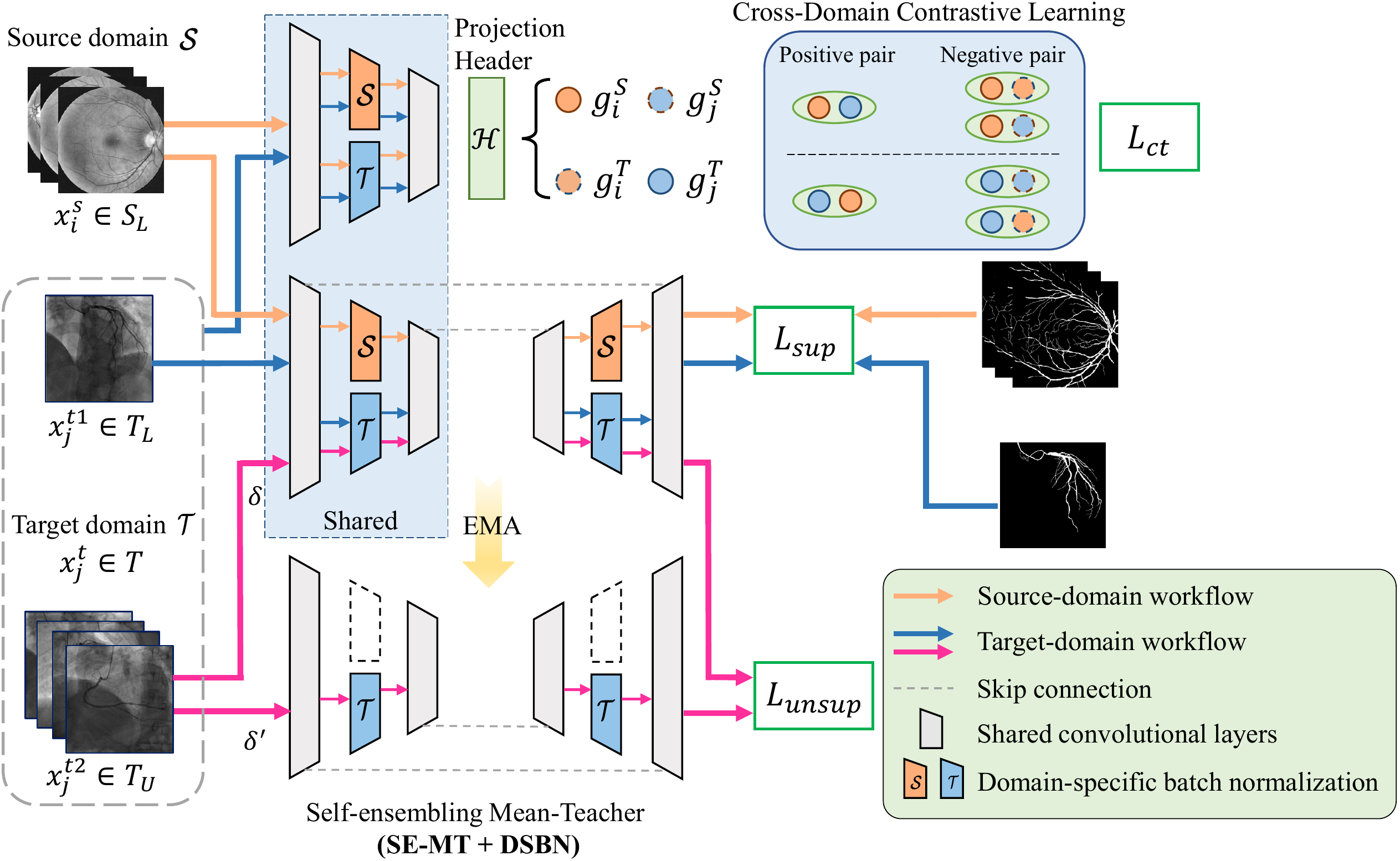}
    \caption{The flowchart of the proposed CS-CADA that consists of three parts: 1) a segmentation network with Domain-Specific Batch Normalization (DSBN); 2) a Self-ensembling Mean-Teacher (SE-MT) architecture; and 3) a cross-domain contrastive learning block.$g^{\mathcal{S}}_{i}$ and $g^{\mathcal{T}}_{i}$ are features of source-domain image $x^{s}_{i}$ normalized by the source- and target-specific BN layers, respectively. $g^{\mathcal{S}}_{j}$ and $g^{\mathcal{T}}_{j}$ are features of target-domain image $x^{t}_{j}$ normalized by the source- and target-specific BN layers, respectively.}
    \label{fig2:model}
\end{figure*}
\section{Methods}
\label{sec3:methods}
Let $\mathcal{S}$ and $\mathcal{T}$ denote the source and the target domains, respectively. We denote the existing annotated source domain dataset as $S_L=\{(x_{i}^{s}, y_{i}^{s})\}^{N_{s}}_{i=1}\}$, and use  $T_L=\{(x_{j}^{t1}, y_{j}^{t1})\}^{N_{t1}}_{j=1}\}$ to denote a small set of annotated images in the target domain. Additionally, another set of unannotated images in the target domain is denoted as $T_U=\{(x_{j}^{t2})\}^{N_{t2}}_{j=1}$, where $N_s$, $N_{t1}$ and $N_{t2}$ are the number of samples in the three datasets,  respectively.
The proposed CS-CADA is depicted in Fig.~\ref{fig2:model}, which consists of three parts: 1) a segmentation network with Domain-Specific
Batch Normalization (DSBN) that learns from $S_L$ and $T_L$ to provide a supervision guidance to bridge the cross-anatomy discrepancy; and 2) a Self-Ensembling Mean-Teacher (SE-MT) that imposes an unsupervised consistency on $T_{U}$ to further enhance data efficiency, and 3) a contrastive learning that encourages the model to capture domain-invariant features. Without loss of generality, we adopt the classical U-Net~\cite{ronneberger2015u} as the backbone for segmentation.

\subsection{Joint Learning with Domain-Specific 
Batch Normalization (DSBN)}
Considering the intensity distribution shift between source and target domains, directly taking $S_L$ and $T_L$ for training without dealing with their difference would have limited performance, since the model will be misguided by statistical variations between domain $\mathcal{S}$ and $\mathcal{T}$ and thus fail to lean general feature representations from the two domains.

To solve the problem, we introduce DSBN block to the network that consists of two types of Batch Normalization (BN) and each of them is in charge of one domain to effectively tackle the inter-domain discrepancy~\cite{chang2019domain}. In our method, DSBN adopts respective BN parameters for domain $\mathcal{S}$ and $\mathcal{T}$ due to the cross-anatomical structure discrepancy. Meanwhile, convolutional kernels are shared across domain $\mathcal{S}$ and $\mathcal{T}$ to learn general representations for the similar anatomical structures. Formally, let $f^{d}\in \mathbb{R}^{N\times H\times W}$ denote feature maps at each channel given by an input from domain $d\in{\{\mathcal{S},\mathcal{T}\}}$. The DSBN normalizes each feature respectively and then applies affine transformation with trainable parameters that are specific to the certain domain $d$, i.e., re-scale parameters $\gamma^{d}$ and bias parameters $\beta^{d}$:
\begin{equation}
\label{eq1:dsbn}
    \hat{f}^{d}=\gamma^{d}\cdot \overline{f}^{d} + \beta^{d},\quad \text { where } \quad \overline{f}^{d}=\frac{f^{d}-\mu^{d}}{\sqrt{(\sigma^{d})^2+\varepsilon}},
\end{equation}
where $\hat{f}^{d}$ is the DSBN output. $\mu^{d}$ and $\sigma^{d}$ are the mean and standard deviation of the features within a mini-batch containing $N$ samples, and $\varepsilon$ is a small number for numeric stability.

Let $\theta_{en}$ and $\theta_{de}$ represent the shared convolutional parameters in the encoder and decoder of the model, respectively. \{$\gamma^{d},\beta^{d}$\} represents a set of trainable parameters in DSBN at domain $d$. Formally, the parameter set for the source domain can be summarized as $\Theta^{\mathcal{S}}=[\theta_{en},\theta_{de},\gamma^{\mathcal{S}},\beta^{\mathcal{S}}]$, and that for the target domain is $\Theta^{\mathcal{T}}=[\theta_{en},\theta_{de},\gamma^{\mathcal{T}},\beta^{\mathcal{T}}]$. DSBN supplies domain-specific variables to handle domain-specific distributions and maps stylistic features to a common space by performing individual feature normalization, which can effectively alleviate the inter-domain discrepancy~\cite{chang2019domain}. During training, DSBN calculates the mean and standard deviation of features for each domain separately, i.e., $\overline{\mu}^{d}$ and $\overline{\sigma}^{d}$. In the test phase, the estimated $\overline{\mu}^{d}$ and $\overline{\sigma}^{d}$ from a moving average in the training stage for each domain are used for whitening input activation.
%The proposed ASBN enables a compact dual-model architecture to deal with individual domain distribution respectively using different parameter sets, i.e., $\Theta^{\mathcal{S}}=[\theta_{en},\theta_{de},\gamma^{\mathcal{S}},\beta^{\mathcal{S}}]$ for the source domain and $\Theta^{\mathcal{T}}=[\theta_{en},\theta_{de},\gamma^{\mathcal{T}},\beta^{\mathcal{T}}]$ for the target domain, which share the convolution parameters $\theta_{en},\theta_{de}$ representing from encoder and decoder of model while use specific BN parameters among each domain, respectively.

Given image-annotation pairs $(x_i^{s},y_i^{s})\in S_L$ from domain $\mathcal{S}$ and $(x_j^{t1},y_j^{t1})\in T_L$ from domain $\mathcal{T}$, we define a supervised loss function to jointly optimize these parameter sets:
\begin{equation}
\label{eq2:loss_sup}
    L_{sup} = \sum_{i=1}^{N_{s}}L_{seg}(p_i^{s},y_i^{s}) + \sum_{j=1}^{N_{t1}}L_{seg}(p_j^{t1},y_j^{t1}),
\end{equation}
where $p_i^{s}=\psi(x_i^{s};\Theta^{\mathcal{S}})$ and $p_j^{t1}=\psi(x_j^{t1};\Theta^{\mathcal{T}})$ are predictions of $x_i^{s}$ and $x_j^{t1}$ using the corresponding parameter sets $\Theta^{\mathcal{S}}$ and $\Theta^{\mathcal{T}}$, respectively. $L_{seg}$ denotes a hybrid segmentation loss that consists of the cross-entropy loss and Dice loss.
\subsection{Self-Ensembling Mean Teacher (SE-MT) with DSBN}
Although the introduced DSBN enables a model to learn from annotated $S_L$ and $T_L$, the small-scale $T_L$ still limits the performance of the model.
To deal with this problem, we employ a Self-Ensembling Mean Teacher (SE-MT) architecture to exploit unannotated images $T_U$ in the target domain.
Specifically, the teacher model $\Theta^{T'}$ is defined as an Exponential Moving Average (EMA) of the student model in the target domain, its parameter in the $k$-th training step is:
\begin{equation}
\label{eq3:ema}
    \Theta^{\mathcal{T}'}_k = \alpha\Theta^{\mathcal{T}'}_{k-1} + (1-\alpha)\Theta^{\mathcal{T}}_k
\end{equation}
where $\alpha\in[0,1]$ is the EMA decay rate~\cite{samuli2017temporal}. The teacher model will guide the student model to give more reliable predictions for unannotated images.
Finally, we define an unsupervised consistency loss ($L_{unsup}$) between the predictions of student model and teacher model for the same input $x_j^{t2}\in T_U$ with different random perturbations $\delta$ and $\delta'$:
\begin{equation}
\label{eq4:loss_unsup}
    L_{unsup} = \sum_{j=1}^{N_{t2}}L_{mse}\left( \psi(x_j^{t2};\Theta^{\mathcal{T}},\delta), \psi(x_j^{t2};\Theta^{\mathcal{T}'},\delta')\right)
\end{equation}
where  $\psi(x_j^{t2};\Theta^{\mathcal{T}},\delta)$ and $\psi(x_j^{t2};\Theta^{\mathcal{T}'},\delta')$ are predictions given by the student and teacher models, respectively. $L_{mse}$ is the mean square error loss. 
\subsection{Cross-Domain Contrastive Learning}
To better deal with the appearance and context shift between the source and target domains, we propose a cross-domain contrastive learning strategy to encourage the model to capture domain-invariant features for the similar anatomical structures while being robust against the different image styles.  

For the output of the encoder with DSBN of the segmentation network, we use a nonlinear projection header $\mathcal{H}$ to obtain a high-level feature representation. For a source domain image $x^{s}_i$, its normalized feature representations 
based on the source-domain BN and target-domain BN are denoted as $g^{\mathcal{S}}_{i}=f(x^{s}_i;\theta_{\mathcal{H}},\theta_{en},\gamma^{\mathcal{S}},\beta^{\mathcal{S}})$ and $g^{\mathcal{T}}_{i}=f(x^{s}_i;\theta_{\mathcal{H}},\theta_{en},\gamma^{\mathcal{T}},\beta^{\mathcal{T}})$, respectively. Correspondingly, for a target domain image $x^{t}_j$, the normalized feature representations based on the source and target domain-specific BNs are denoted as $g^\mathcal{S}_j=f(x^{t}_j;\theta_{\mathcal{H}},\theta_{en},\gamma^{\mathcal{S}},\beta^{\mathcal{S}})$ and $g^{\mathcal{T}}_j=f(x^{t}_j;\theta_{\mathcal{H}},\theta_{en},\gamma^{\mathcal{T}},\beta^{\mathcal{T}})$, respectively.

As the DSBN aims to normalize the feature maps of two domains respectively so that they are mapped to a common feature distribution space to alleviate the domain gap. Given a pair of images from the source and target domains $x^{s}_i$ and $x^{t}_j$, $g^{\mathcal{S}}_{i}$ and $g^{\mathcal{T}}_{j}$ are both normalized features in the common distribution space based on their specific BNs. They should be similar to each other, and thus be set as a positive pair. For the pair of ($g^{\mathcal{S}}_{i}$, $g^{\mathcal{T}}_{i}$), they are resulted from the same feature of a source-domain image $x^{s}_{i}$ normalized by two different BNs, they should be different from each other due to the different styles associated with the two BNs. Therefore, they are set as a negative pair.

%Because domain-specific batch normalization only regular a specific domain feature distribution, and specific batch normalization between source and target can map two different distributed feature on a common Reproducing Kernel Hilbert Space (RKHS) and train the segmentation task on a model. To conduct cross-domain contrastive learning, we defined one positive sample pair. 
%Following Wang et. al~\cite{wang2021dense} suggestion that more negative pairs benefiting for model performance, we designed two negative sample pairs since each pair simulating different domain distribution under a certain content.
%All feature representations paired for contrastive learning are from the output of the encoder, since it contains more high dimensional feature which can be seemed as the basis of domain-invariant feature.

%For collected training samples in contrastive learning, we choose $g^{\mathcal{S}}_{i}$ and $g^{\mathcal{T}}_j$ as positive pair, Meanwhile, for negative pairs, we define four types of negative samples including $g^{\mathcal{S}}_{i}$ accompanied with $g^{\mathcal{T}}_{i}$, $g^{\mathcal{S}}_{i}$ accompanied with $g^{\mathcal{S}}_{j}$, as well as $g^{\mathcal{T}}_j$ accompanied with $g^{\mathcal{S}}_j$, and $g^{\mathcal{T}}_j$ accompanied with $g^{\mathcal{T}}_i$. 
Following the standard formula of self-supervised contrastive loss~\cite{chen2020simple,wang2021dense}, we define a source to target domain contrastive loss as:
\begin{equation}
\label{eq5_1:loss_contrast_s2t}
\begin{aligned}
    L_{ct}^{s2t} &= -\mathbb{E}_{\scriptscriptstyle x^{s}_{i},x^{t}_j\sim\mathcal{S},\mathcal{T}}\Big(\\
    & log\big(\frac{e^{sim(g^\mathcal{S}_{i}, g^\mathcal{T}_{j})/\tau}}{e^{sim(g^\mathcal{S}_{i},g^\mathcal{T}_{j})/\tau}+\sum_{g\in\mathcal{N}_i} e^{sim(g^\mathcal{S}_{i}, g)/\tau}} \big)\Big)
\end{aligned}
\end{equation}
where $\mathcal{N}_i = \{g^\mathcal{T}_{i}, g^\mathcal{S}_{j}\}$ are the negative counterparts of $g^\mathcal{S}_{i}$. The $sim(\cdot,\cdot)$ is the cosine similarity between two representations, and $\tau=0.1$ is the temperature scaling parameter. Correspondingly, for $g^\mathcal{T}_{j}$, we define a target to source domain contrastive loss as:
\begin{equation}
\label{eq5_2:loss_contrast_t2s}
\begin{aligned}
    L_{ct}^{t2s} &= -\mathbb{E}_{\scriptscriptstyle x^{s}_{i},x^{t}_j\sim\mathcal{S},\mathcal{T}}\Big(\\
    & log\big(\frac{e^{sim(g^\mathcal{T}_{j}, g^\mathcal{S}_{i})/\tau}}{e^{sim(g^\mathcal{T}_{j},g^\mathcal{S}_{i})/\tau}+\sum_{g\in\mathcal{N}_j} e^{sim(g^\mathcal{T}_{j}, g)/\tau}} \big)\Big)
\end{aligned}
\end{equation}
%where $(D^\mathcal{\alpha}, D^\mathcal{\beta}) \in \{(g^\mathcal{T}_{j},g^\mathcal{S}_{j}),(g^\mathcal{T}_{j},g^\mathcal{T}_{i})\}$. 
where $\mathcal{N}_j = \{g^\mathcal{S}_{j}, g^\mathcal{T}_{i}\}$ are the negative counterparts of $g^\mathcal{T}_{j}$.
Finally, the cross-domain contrastive loss is formulated as:
\begin{equation}
\label{eq5:loss_contrast}
    L_{ct} = \frac{1}{2}(L_{ct}^{s2t}+L_{ct}^{t2s})
\end{equation}
\subsection{Overall Training Loss}
The overall loss for training is a combination of the supervised loss $L_{sup}$ in Eq.~\ref{eq2:loss_sup}, the unsupervised consistency loss $L_{unsup}$ in Eq.~\ref{eq4:loss_unsup} and the contrastive learning loss $L_{ct}$ in Eq.~\ref{eq5:loss_contrast}. It is formulated as:
\begin{equation}
\label{eq5:loss_overall}
    L = L_{sup} + \lambda_{1}\cdot L_{unsup} + \lambda_{2}\cdot L_{ct}
\end{equation}
where $\lambda_1,\lambda_2$ act as trade-off parameters. Once the training process is completed, segmentation results for the target domain images are obtained by a forward propagation on the student model with parameter set $\Theta^{\mathcal{T}}$.
\section{Experimental Results}
\label{sec4:experiments}
%We designed and collected two groups of dataset to conduct cross-anatomy domain adaptation for similar anatomical structure segmentation from different organs. 1). Digital Retinal Images for Vessel Extraction (DRIVE) and X-ray Angiograms (XAs) for coronary artery segmentation which representing vascular structure segmentation task; 2). Retinal Fundus Glaucoma Challenge (REFUGE) for optic cup and disk segmentation and Multi-Sequence Cardiac MR Segmentation (MS-CMRSeg) Challenge for the left ventricle blood cavity (LV), and the myocardium of the left ventricle (Myo) segmentation which representing a more challenge task setting, referring the cyclic structure segmentation.
\begin{table*}[t]
    \caption{Quantitative comparison between our CS-CADA and state-of-the-art methods for cardiac artery segmentation. Only $10\%$ of training images in the target domain are annotated. Fine-tuning (last) means only updating the last convolutional block of the decoder. Fine-tuning (all) means updating all the parameters of the model.}
    % \vspace{1mm}
    \footnotesize
    \centering
    \scalebox{1.15}{\begin{tabular}{ll|lll|lll}
    \hline
    \hline
    \multicolumn{2}{c|}{\multirow{2}{*}{Methods}} & \multicolumn{3}{c|}{Training set} & \multirow{2}{*}{Recall (\%)} & \multirow{2}{*}{Precision (\%)} & \multirow{2}{*}{Dice (\%)} \\
    \cline{3-5}
    & & $S_{L}$ & $T_{U}$ & $T_{L}$ &  &  & \\
    \hline
    \multicolumn{2}{l|}{Baseline (source)} & \ding{51} & & & 34.68$\pm$6.22 & 40.84$\pm$2.79 & 37.31$\pm$4.90 \\
    \multicolumn{2}{l|}{Baseline (target)} & & & \ding{51} & 75.12$\pm$4.83 & 65.87$\pm$8.47 & 69.81$\pm$5.35 \\
    \hline
    \multirow{3}{*}{UDA} 
    & ADDA~\cite{tzeng2017adversarial} & \ding{51} & \ding{51} & & 43.13$\pm$15.08 & 39.49$\pm$6.49 & 40.24$\pm$9.34 \\
    & SIFA~\cite{chen2019synergistic} & \ding{51} & \ding{51} & & 62.62$\pm$10.15 & 46.66$\pm$7.43 & 52.44$\pm$4.62 \\
    & SC-GAN~\cite{yu2019annotation} & \ding{51} & \ding{51} &  & 80.87$\pm$7.69 & 57.88$\pm$8.73 & 66.92$\pm$6.83 \\
    \hline
    \multirow{4}{*}{SDA} & Fine-tuning(last)~\cite{Tajbakhsh2016fine} & \ding{51} & & \ding{51} & 56.76$\pm$7.12 & 52.55$\pm$14.63 & 53.75$\pm$11.27 \\ 
    & Fine-tuning(all) & \ding{51} & & \ding{51} & 77.44$\pm$4.46 & 73.81$\pm$2.75 & 75.50$\pm$2.72 \\
    & Joint Training & \ding{51} & & \ding{51} & 81.14$\pm$6.75 & 62.66$\pm$9.35 & 70.37$\pm$7.37 \\
    & X-shape~\cite{valindria2018multi} & \ding{51} & & \ding{51} & 83.10$\pm$4.11 & 65.11$\pm$10.92 & 72.49$\pm$7.69\\
    & DSBN~\cite{chang2019domain} & \ding{51} & & \ding{51} & 80.66$\pm$4.58 & 69.12$\pm$9.53 & 74.12$\pm$6.62\\
    \hline
    \multirow{3}{*}{SSL} & SE-MT~\cite{french2018selfensembling} & & \ding{51} & \ding{51} & 82.55$\pm$3.89 & 72.14$\pm$8.40 & 76.70$\pm$5.43 \\
    & UA-MT~\cite{yu2019uncertainty} & & \ding{51} & \ding{51} & 82.40$\pm$2.88 & 69.13$\pm$4.35 & 75.13$\pm$3.35  \\
    & CPS~\cite{chen2021semi} & & \ding{51} & \ding{51} & 82.13$\pm$3.36 & 72.02$\pm$3.80 & 76.69$\pm$3.10  \\
    \hline
    SSDA & Dual-T~\cite{li2020dual} & \ding{51} & \ding{51} & \ding{51} & 82.63$\pm$2.89 & 70.33$\pm$3.14 & 75.94$\pm$2.53 \\ 
    \hline
    % \multicolumn{2}{l|}{SS-CADA (ours)} & \ding{51} & \ding{51} & \ding{51} & 82.76$\pm$3.95 & 75.12$\pm$6.59 & 78.51$\pm$4.60\\
    % \hline
    \multicolumn{2}{l|}{\textbf{CS-CADA (ours)}} & \ding{51} & \ding{51} & \ding{51} & \textbf{83.13$\pm$3.87} & \textbf{77.32$\pm$6.55} & \textbf{79.28$\pm$3.67} \\
    \hline
    \hline
    \end{tabular}}
    \label{tab1:vessel_sota}
\end{table*}
\begin{figure*}[t]
    \centerline{\includegraphics[width=7in]{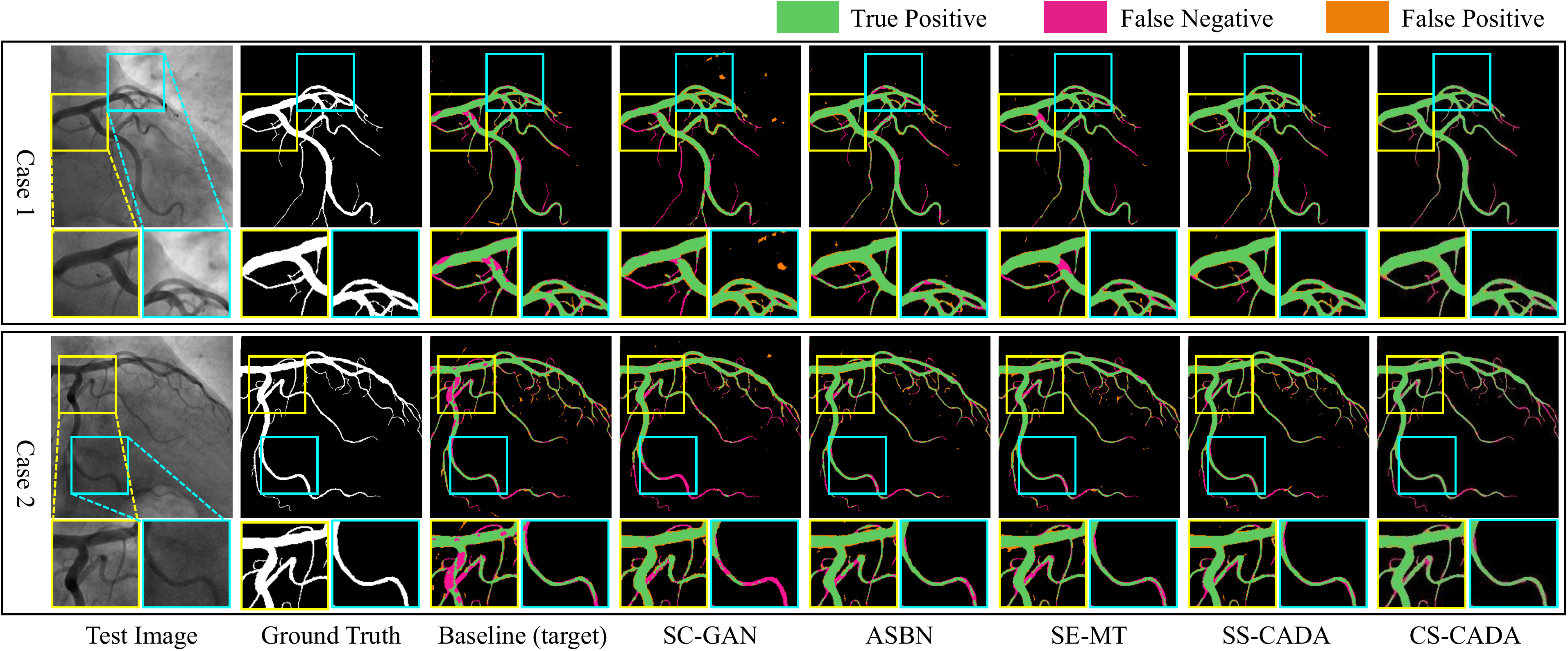}}
    \caption{Visual comparison of different methods for coronary artery segmentation from XAs. The true positives, false negatives and false positives are colored in green, red and orange, respectively. The zoomed views are appended below each case to highlight the segmentation details.}
    \label{fig4:vessel_seg}
\end{figure*}
\subsection{Implementation Details}
For experiments, we used the U-Net~\cite{ronneberger2015u} as segmentation network in the mean-teacher backbone, and extended it with DSBN for domain-specific feature normalization. The EMA decay rate was empirically set to 0.99, and a time-dependent Gaussian warming up function $\alpha(k)=0.1e^{\left(-5\left(1-k / k_{max}\right)^{2}\right)}$ was used to dynamically update hyper parameter $\alpha$, where $k$ denotes the current training iteration and $k_{max}$ is the last iteration. Each model was trained by the Adam optimizer with max iteration $20,000$ and initial learning rate $5e^{-4}$ that was decayed exponentially with power $0.95$. For training, the loss function weights are set as $\lambda_1=1$ and $\lambda_2=0.1$, respectively. The batch size was 8 and 12 for vascular structure segmentation and circular structure segmentation, respectively. And in each batch, half of the images are labeled and half of them are unlabeled. All experiments were implemented by Pytorch with one NVIDIA Geforce GTX 1080 Ti GPU.

Our CS-CADA was validated with two applications: 1) adapting a model from digital retinal images with annotated vessels to segment coronary artery from XAs, and 2) adapting a model from retinal fundus images with annotated optic cup and disk to segment left ventricle blood cavity and myocardium from cardiac MRI. Each of two domain images were processed by contrast limited adaptive histogram equalization and gamma correction. 
\subsection{Vascular Structure Segmentation}
\label{subsec:vascular_seg}
\subsubsection{Datasets} The first scenario is modal adaptation for vascular structure segmentation. We used the DRIVE~\cite{DRIVE} dataset as the source domain dataset $S_L$. It contains 40  Fundus Images (FI) of retinal vessels acquired from a Canon CR5 nonmydriatic 3CCD camera at 45$^\circ$ field of view. In addition, we collected 191 XAs of 30 patients using a Philips UNIQ FD10 C-arm system with coronary arteries as the target domain. An expert radiologist randomly annotated 14 XAs of 2 patients to serve as $T_L$, and we took 121 XAs of 19 patients without annotations as $T_U$. The other 20 XAs of 3 patients and 36 XAs of 6 patients were used for validation and testing, respectively. For FIs, we only used the green channel, and  all FIs and XAs were resized to $512\times512$. We used random horizontal and vertical flipping, random cropping and resize with an output size of $400\times400$ for data augmentation. The image intensity was normalized to [0, 1]. The batch size was $16$, where $8$, $4$ and $4$ images were from $S_L$, $T_L$ and $T_U$, respectively. The first row of Fig.~\ref{fig1:domain_shift} shows a comparison of samples from two domains and their corresponding histograms.
\begin{table}[t]
    \caption{Quantitative ablation study of our CS-CADA for coronary artery segmentation. SE-MT is semi-supervised learning only using $T_L$ and $T_U$. DSBN means using the Domain-specific batch normalization for learning from $S_L$ and $T_L$. SS-CADA means learning from $S_L$, $T_L$ and $T_U$ at the same time based on our introduced DSBN and SE-MT, without using contrastive learning.}
    % \vspace{1mm}
    \small
    \centering
    \scalebox{1}{\begin{tabular}{l|lll}
    \hline
    \hline
    Methods & Recall (\%) & Precision (\%) & Dice (\%) \\ \hline
    Baseline (target) & 75.12$\pm$4.83 & 65.87$\pm$8.47 & 69.81$\pm$5.35 \\
    SE-MT & 82.55$\pm$3.89 & 72.14$\pm$8.40 & 76.70$\pm$5.43 \\
    DSBN & 80.66$\pm$4.58 & 69.12$\pm$9.53 & 74.12$\pm$6.62 \\
    SS-CADA~\cite{zhang2021sscada} & \textbf{83.27$\pm$3.95} & 75.12$\pm$6.59 & 78.84$\pm$4.60 \\
    \textbf{CS-CADA (ours)} & 83.13$\pm$3.87 & \textbf{77.32$\pm$6.55} & \textbf{79.28$\pm$3.67} \\ 
    %Upper-bound & 81.45$\pm$3.88 & 84.89$\pm$3.24 & 83.50$\pm$3.07 \\
    \hline
    \hline
    \end{tabular}}
    \label{tab2:vessel_ablation}
\end{table}
\subsubsection{Comparison with Existing Methods}
\label{subsubsec:vel_sota}
To demonstrate the effectiveness of CS-CADA for solving cross-anatomy domain shift, we compared it with the baseline in two settings: 1) Using only $S_L$: A standard U-Net only learns from the source domain images, which is denoted as Baseline (source); 2) Using only $T_L$: A standard U-Net learns only from the labeled target domain images, which is denoted as Baseline (target). We also compared it with several state-of-the-art methods in four categories: 
1) UDA methods that use $S_L$ and unlabeled $T_U$: We investigated three UDA methods including ADDA~\cite{tzeng2017adversarial}, %that first outlined a generalized framework for adversarial adaptation, 
SIFA~\cite{chen2019synergistic}, %that presented synergistic fusion of adaptations, 
and SC-GAN~\cite{yu2019annotation}; %that incorporated shape prior knowledge through predicted pseudo label from target domain; 
2) Supervised Domain Adaptation (SDA)~\cite{Tajbakhsh2016fine} methods that use $S_L$ and $T_L$ for training. We considered four methods, including Joint Training that takes $S_L \cup T_L$ as a single uniform training set, Fine-tuning~\cite{Tajbakhsh2016fine} where the segmentation model is pre-trained on $S_L$ and fine-tuned with $T_L$. Here, we consider two types of fine-tuning strategies: fine-tuning (last)  means only updating parameters in last convolutional block of the decoder, and fine-tuning (all) means updating the whole set of parameters of the model. Both fine-tuning methods used 2000 iterations. X-shape~\cite{valindria2018multi} that separately learns from $S_L$ and $T_L$, 
and DSBN~\cite{chang2019domain} that uses domain-specific batch normalization for joint training; 
3) SSL methods that use $T_L$ and $T_U$ for training, and we considered three state-of-the-art methods including SE-MT~\cite{french2018selfensembling}, UA-MT~\cite{yu2019uncertainty} and Cross Pseudo Supervision (CPS)~\cite{chen2021semi};
and 4) Semi-Supervised Domain Adaptation (SSDA) method, i.e., Dual-teacher (Dual-T)~\cite{li2020dual}. %constrains domain shift by introducing intra-domain and inter-domain consistency.
All the compared methods were quantitatively evaluated with Recall, Precision and Dice score.

Comprehensive experimental results are shown in Table~\ref{tab1:vessel_sota}. It presented that Baseline (source) only achieved an average Dice of $37.31\%$, showing the large domain gap between FIs and XAs. Baseline (target) only obtained average Dice at $69.81\%$, indicating that using a small set of labeled XAs cannot lead to accurate results. 
The UDA methods outperformed Baseline (source), and SC-GAN~\cite{yu2019annotation} was better than ADDA~\cite{tzeng2017adversarial} and SIFA~\cite{chen2019synergistic}, but its performance is worse than baseline (target) due to the large domain gap and they do not use supervision from the target domain. 
For SDA methods, Fine-tuning (all)~\cite{Tajbakhsh2016fine} achieved the highest Dice score of $75.50\%$. Joint Training, X-shape~\cite{valindria2018multi} and DSBN~\cite{chang2019domain} based methods achieved better results than fine tuning the only last block of the decoder and UDA methods, demonstrating that small set of labeled XAs can provide effective supervision for bridging the cross-anatomy domain shift, in which the DSBN got the highest Dice ($74.12\%$) among them. The SSL methods generally performed better than the SDA methods, showing the usefulness of unannotated images in the target domain. However, all these methods were inferior to our proposed CS-CADA that obtained an average Dice of $79.28\%$, which is a large improvement from $75.94\%$ obtained by existing SSDA method Dual-T~\cite{li2020dual}.  %SE-MT and CPS only use the knowledge specific to coronary arteries in XAs yet without transferable knowledge from retinal vessels in FIs to XAs leading to around $2.58\%$ decrease in Dice compared with CS-CADA. %Further uncertainty based UA-MT obtains even lower Dice ($75.13\%$) in cross-anatomy task. 
%For SSDA method, Dual-teacher method can get the Dice of $75.94\%$ but can gain a lower standard deviation than SE-MT. In comparison, our proposed CS-CADA achieve higher performance than above mentioned methods, in which it gains promising performance achieving $79.28\%$ in Dice score when anatomy-specific batch normalization and cross-domain contrastive learning strategy were exploited  under the semi-supervised backbone to tackle cross-anatomy domain shift. The efficiency of our proposed CS-CADA can also be proved in its 
Our method also has higher Precision and Recall than the other methods as shown in Table~\ref{tab1:vessel_sota}.

Fig.~\ref{fig4:vessel_seg} is visual comparison between different methods in two cases. We selected SC-GAN~\cite{yu2019annotation}, DSBN~\cite{chang2019domain} and SE-MT~\cite{french2018selfensembling} that are the best UDA, SDA and SSL method shown in Table~\ref{tab1:vessel_sota}, respectively. SC-GAN exhibits obvious false negatives for thin terminal vessels, and false positives scattered in the background, which demonstrates that UDA methods designed for different domains with the same anatomical structure are not suitable for cross-anatomy adaptation. 
Compared with SC-GAN, DSBN obtained less false positives, but the connectivity of coronary arteries is not well preserved. SE-MT  restored more details in the results, but the false negative region is visually obvious. In contrast, our CS-CADA obtained accurate and detailed results with very small amount of false negatives and false positives even for the thin terminals. % Based on it, given limited labeled XAs, only our proposed SS-CADA and CS-CADA enables accurate extraction of coronary arteries with the best connectivity and the least false positives. Meanwhile, CS-CADA owns sensitivity on structural details, restoring as much as possible detail in bifurcation points and thin vessels.
\begin{figure}[t]
    \centerline{\includegraphics[width=3in]{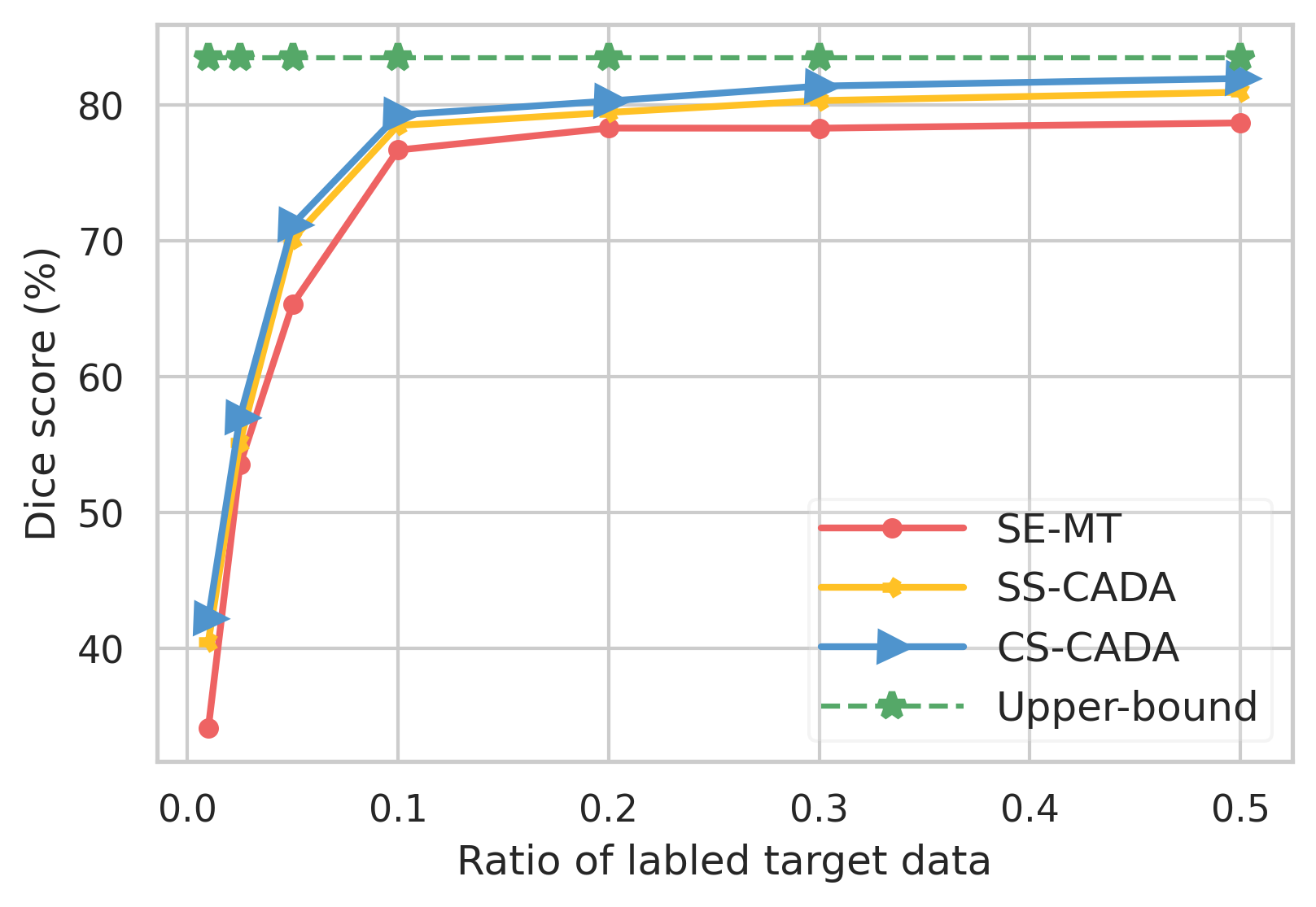}}
    \caption{Average Dice under different ratios of annotated XAs for coronary artery segmentation. The green dotted line is the upper bound of fully supervised learning.}
    \label{fig5:vel_ablations}
\end{figure}
\begin{table*}[t]
    \caption{Quantitative comparison between our CS-CADA and state-of-the-art methods for LV and Myo segmentation. Only $10\%$ of training images in the target domain are annotated. Fine-tuning (last) means only updating the last convolutional block of the decoder. Fine-tuning (all) means updating all the parameters of the model.}
    % \vspace{1mm}
    \footnotesize
    \centering
    \scalebox{1.05}{\begin{tabular}{p{0.45cm}p{2.3cm}|p{0.2cm}p{0.2cm}p{0.2cm}|p{1.32cm}p{1.32cm}|p{1.32cm}p{1.32cm}|p{1.3cm}p{1.32cm}}%{ll|lll|ll|ll|l}
    \hline
    \hline
    \multicolumn{2}{c|}{\multirow{3}{*}{Methods}} & \multicolumn{3}{c|}{Training set} & \multicolumn{2}{c|}{LV} & \multicolumn{2}{c|}{Myo} & 
    \multicolumn{2}{c}{Average} \\
    \cline{3-11}
    & & $S_{L}$ & $T_{U}$ & $T_{L}$ & Dice (\%) & ASSD (mm) & Dice (\%) & ASSD (mm) & Dice (\%) & ASSD (mm)  \\
    \hline
    \multicolumn{2}{l|}{Baseline (source)} & \ding{51} & & & 9.17$\pm$10.03 & 49.54$\pm$23.66 & 11.07$\pm$2.33 & 35.71$\pm$2.56 & 10.12$\pm$5.45 & 42.63$\pm$26.48 \\
    \multicolumn{2}{l|}{Baseline (target)} & & & \ding{51} & 77.87$\pm$6.63 & 6.96$\pm$3.35 & 54.56$\pm$7.93 & 8.37$\pm$2.98 & 66.21$\pm$4.66 & 7.67$\pm$3.67 \\
    \hline
    \multirow{3}{*}{UDA}
    & ADDA~\cite{tzeng2017adversarial} & \ding{51} & \ding{51} & & 40.59$\pm$23.30 & 8.57$\pm$9.51 & 20.32$\pm$11.60 & 12.98$\pm$4.85 & 30.45$\pm$16.38 & 10.78$\pm$7.94 \\
    & CycleGAN~\cite{zhu2017unpaired} & \ding{51} & \ding{51} & & 51.52$\pm$8.58 & 6.98$\pm$2.96 & 26.62$\pm$3.74 & 8.64$\pm$2.43 & 39.07$\pm$4.87 & 7.81$\pm$3.13 \\
    & SIFA~\cite{chen2019synergistic} & \ding{51} & \ding{51} & & 78.81$\pm$11.14 & 5.97$\pm$2.74 & 41.13$\pm$7.08 & 9.44$\pm$1.94 & 59.97$\pm$7.72 & 7.71$\pm$2.95 \\
    \hline
    \multirow{4}{*}{SDA} & Fine-tuning(last)\cite{Tajbakhsh2016fine} & \ding{51} & & \ding{51} & 45.43$\pm$11.07 & 25.22$\pm$6.80 & 26.06$\pm$9.54 & 20.60$\pm$6.09 & 35.75$\pm$9.68 & 22.91$\pm$6.85 \\ 
    & Fine-tuning(all) & \ding{51} & & \ding{51} & 85.51$\pm$5.37 & 8.79$\pm$6.02 & 71.74$\pm$4.89 & 6.00$\pm$3.48 & 78.63$\pm$4.44 & 7.40$\pm$5.11 \\
    & Joint Training & \ding{51} & & \ding{51} & 75.70$\pm$11.03 & 9.64$\pm$7.70 & 56.95$\pm$9.41 & 8.06$\pm$4.54 & 66.32$\pm$9.88 & 8.85$\pm$7.56 \\
    & X-shape~\cite{valindria2018multi} & \ding{51} & & \ding{51} & 82.92$\pm$9.38 & 3.39$\pm$2.35 & 67.91$\pm$6.44 & 4.44$\pm$2.30 & 75.41$\pm$7.39 & 3.92$\pm$3.27 \\
    & DSBN~\cite{chang2019domain} & \ding{51} & & \ding{51} & 84.97$\pm$5.75 & 5.08$\pm$2.98 & 70.63$\pm$6.31 & 7.46$\pm$2.88 & 77.80$\pm$5.47 & 6.27$\pm$2.94 \\
    \hline
    \multirow{3}{*}{SSL} & SE-MT~\cite{french2018selfensembling} & & \ding{51} & \ding{51} & 82.10$\pm$5.39 & 5.29$\pm$3.21 & 69.18$\pm$4.84 & 5.46$\pm$2.63 & 75.65$\pm$4.66 & 5.38$\pm$3.27 \\
    & UA-MT~\cite{yu2019uncertainty} & & \ding{51} & \ding{51} & 81.56$\pm$6.13 & 5.40$\pm$3.74 & 71.95$\pm$5.49 & 5.40$\pm$1.51 & 76.76$\pm$3.74 & 4.40$\pm$3.03 \\
    & CPS~\cite{chen2021semi} & & \ding{51} & \ding{51} & 82.53$\pm$10.42 & 4.18$\pm$3.25 & 67.39$\pm$7.61 & 4.70$\pm$2.44 & 74.96$\pm$8.54 & 4.44$\pm$2.89\\
    \hline
    SSDA & Dual-T~\cite{li2020dual} & \ding{51} & \ding{51} & \ding{51} & 81.81$\pm$8.90 & 4.57$\pm$3.14 & 70.62$\pm$5.45 & 5.32$\pm$2.25 & 76.22$\pm$6.74 & 4.94$\pm$2.76 \\ 
    \hline
    % \multicolumn{2}{l|}{SS-CADA (ours)} & \ding{51} & \ding{51} & \ding{51} & 86.76$\pm$5.88 & 3.37$\pm$2.17 & 71.74$\pm$6.88 & 4.30$\pm$2.19 & 79.25$\pm$5.94 \\
    % \hline
    \multicolumn{2}{l|}{\textbf{CS-CADA (ours)}} & \ding{51} & \ding{51} & \ding{51} & \textbf{87.02$\pm$4.77} & \textbf{3.15$\pm$2.20} & \textbf{74.85$\pm$5.90} & \textbf{3.37$\pm$2.04} & \textbf{80.80$\pm$4.94} & \textbf{3.26$\pm$2.15} \\
    \hline
    \hline
    \end{tabular}}
    \label{tab3:cardiac_sota}
\end{table*}
\subsubsection{Ablations Study}
\label{subsubsec:vel_abla}
For ablation study, we compared our CS-CADA with three variants: 1) SE-MT that does not use the source domain images therefore without DSBN and $L_{ct}$, 2) DSBN that only learns from $S_L$ and $T_L$ based on shared convolution with domain-specific batch normalization, and 3) a combination of SE-MT and DSBN, without using $L_{ct}$, which is denoted as SS-CADA~\cite{zhang2021sscada}. The results in Table~\ref{tab2:vessel_ablation} show that SS-CADA combining $S_L$, $T_L$ and $T_U$ performed better than only learning from $T_L + T_U$ (SE-MT) and $S_L + T_L$ (DSBN). Finally, the comparison between SS-CADA and CS-CADA highlights the contribution of our proposed cross-domain contrastive learning strategy. The results show that by leveraging the existing data from another domain and the unannotated images in the target domain, our method improved the average Dice by $13.56\%$ (i.e., from $69.81\%$ to $79.28\%$). We conducted a statistical significance evaluation based on paired t-test between SS-CADA and CS-CADA. The p-value of Recall was 0.67 $>$ 0.05 meaning there is no significant difference. Although the p-value of Dice score was 0.076 $>$ 0.05, the p-value of Precision was 0.007 $<$ 0.05, showing the significant improvement from SS-CADA to CS-CADA. Visual comparison between SS-CADA and CS-CADA are shown in the last two columns of Fig.~\ref{fig4:vessel_seg}.

To investigate the annotation efficiency for our semi-supervised method with additional source domain images for training, we experimented with different ratios of labeled images in the target domain, i.e., $1\%$, $3\%$, $5\%$, $10\%$, $30\%$ and $50\%$ respectively. We compared our method with SE-MT and SS-CADA that are better than the other variants according to Table~\ref{tab2:vessel_ablation}.  Fig.~\ref{fig5:vel_ablations} shows that all these methods have a poor performance when the annotation ratio is below 10\%, and the performance is much improved with a higher annotation ratio. Our proposed CS-CADA consistently outperformed SS-CADA and SE-MT with different annotation ratios. When the annotation ratio is 0.3 to 0.5, the performance of our CS-CADA is close to that of fully supervised learning, showing effectiveness of our method for reducing the annotation requirement in the target domain. 
%The large performance differences are presented when fed less $10\%$ ratio of labeled XAs.
Meanwhile, the performance gap between SE-MT and CS-CADA is larger when the annotation ratio is smaller,  which reveals that our method have more advantages than existing methods when the amount of annotations are very limited (e.g., only less than 10$\%$ of the annotations are available). %Two strategies both can achieve state-of-the-art performance on their effectual range forming a situation of complementary advantages.
\label{subsec:circular_seg}
\begin{figure*}[t]
    \centerline{\includegraphics[width=6.8in]{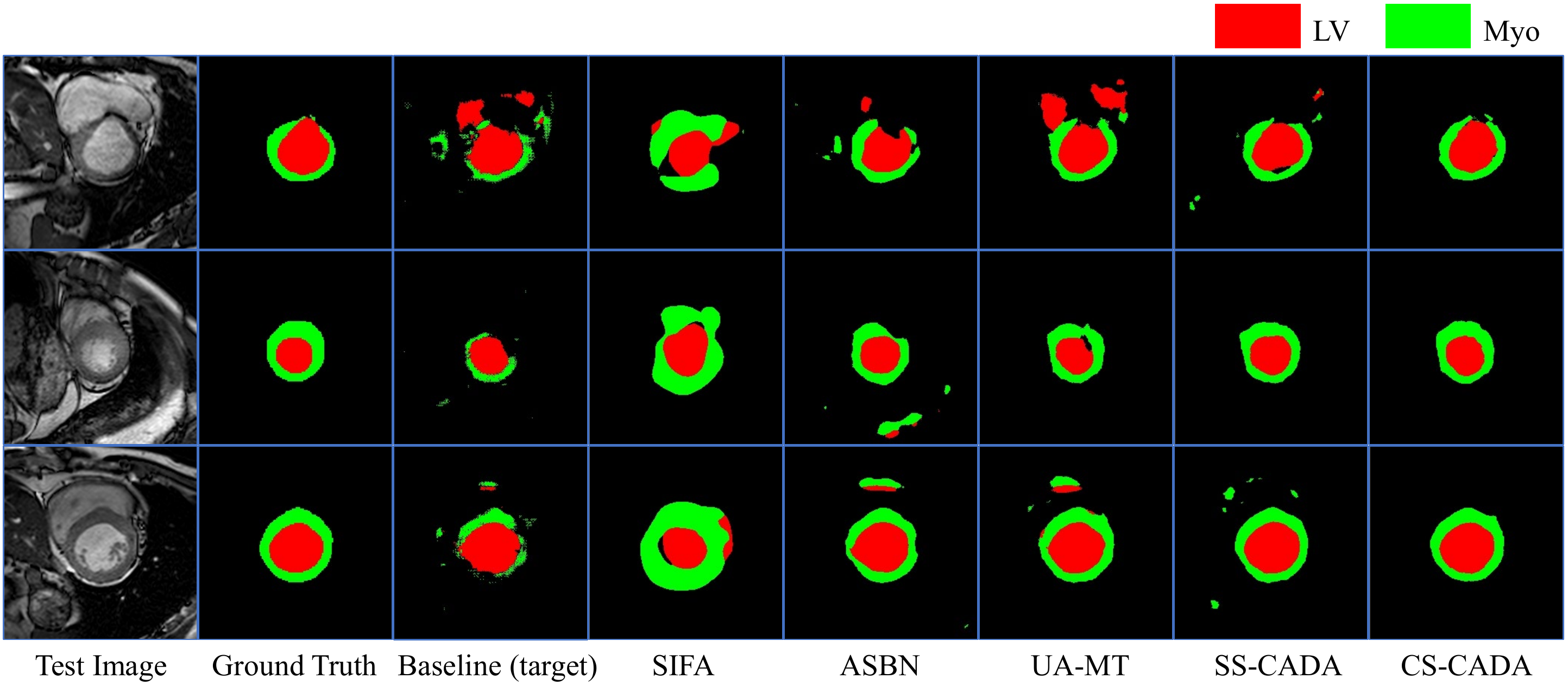}}
    \caption{Visual comparison of different methods for LV and Myo segmentation. Red and green regions denote the LV and Myo, respectively.}
    \label{fig6:cardiac_seg}
\end{figure*}
\subsection{Circular Structure Segmentation}
\subsubsection{Datasets}
We then applied our method to Left Ventricle (LV) and left ventricular Myocardium (Myo) segmentation from MRI. As the LV and Myo are circular structures that have similar shapes with optic disc and cup, we use the REFUGE~\cite{orlando2020refuge} dataset with annotated optic disc and cup as source domain images. It contains 400 retinal fundus images  with a size of $2124\times2056$, acquired by a Zeiss Visucam 500 camera. %400 labeled (testing dataset), and 400 additional unlabeled (unsupervised training dataset) target domain retinal fundus images with size $1634\times1634$ collected by a Canon CR-2 camera.
As for glaucoma patients, the shapes of optic disc and cup may be abnormal, we only used the 360 non-glaucoma images in that dataset as our $S_L$. For preprocessing, we first cropped the image with a patch size of $640\times640$ centered on the optic disk and then resized it to $256\times256$.
Meanwhile, we utilized the Multi-Sequence Cardiac MR segmentation challenge (MS-CMRSeg) dataset~\cite{zhuang2018multivariate} as target domain images that consists of 45 multi-sequence CMR images from patients with cardiomyopathy. Each patient was scanned with LGE, T2-weighted and bSSFP sequences respectively, and we only used the bSSFP sequence for the segmentation task. The dataset has 412 2D slices in total, and we randomly split them into 282 slices of 32 patients for training, 36 slices of 4 patients for validation and 94 slices of 9 patients for testing. For the 282 training slices, 28 slices were used as $T_L$ and the others were used as $T_U$. For preprocessing, the image was firstly cropped with a patch size of $160\times160$ centered on the target region and then resized to $256\times256$. All dataset were normalized the intensity to [0, 1] in training. The batch size was $24$ where $12$, $6$ and $6$ images were from $S_L$, $T_L$ and $T_U$, respectively. The second row of Fig.~\ref{fig1:domain_shift} shows two examples of images in the source and target domains and their histograms.

\begin{table*}[t]
    \caption{Quantitative ablation study of our CS-CADA for LV and Myo segmentation. SE-MT is semi-supervised learning only using $T_L$ and $T_U$. DSBN means using the anatomy-specific batch normalization for learning from $S_L$ and $T_L$. SS-CADA means learning from $S_L$, $T_L$ and $T_U$ at the same time using our proposed DSBN and SE-MT, without using contrastive learning.}
    % \vspace{1mm}
    \small
    \centering
    \scalebox{1}{\begin{tabular}{l|ll|ll|ll}
    \hline
    \hline
   \multirow{3}{*}{Methods} & \multicolumn{2}{c|}{LV} & \multicolumn{2}{c|}{Myo} & \multicolumn{2}{c}{Average} \\ \cline{2-7}
    & Dice ({\%}) & ASSD ({mm}) & Dice ({\%}) & ASSD ({mm}) & 
    Dice ({\%}) &  ASSD ({mm}) \\ \hline
    Baseline (target) & 77.87$\pm$6.63 & 6.96$\pm$3.35 & 54.56$\pm$7.93 & 8.37$\pm$2.98 & 66.21$\pm$4.66 & 7.67$\pm$3.67 \\
    SE-MT & 82.10$\pm$5.39 & 5.29$\pm$3.21 & 69.18$\pm$4.84 & 5.46$\pm$2.63 & 75.65$\pm$4.66 & 5.38$\pm$3.27 \\
    DSBN & 84.97$\pm$5.75 & 5.08$\pm$2.98 & 70.63$\pm$6.31 & 7.46$\pm$2.88 & 77.80$\pm$5.47 & 6.27$\pm$2.94 \\
    SS-CADA~\cite{zhang2021sscada} & 86.76$\pm$5.88 & 3.37$\pm$2.17 & 71.74$\pm$6.88 & 4.30$\pm$2.19 & 79.25$\pm$5.94 & 3.84$\pm$2.21 \\
    \textbf{CS-CADA (ours)} & \textbf{87.02$\pm$4.77} & \textbf{3.15$\pm$2.20} & \textbf{74.85$\pm$5.90} & \textbf{3.37$\pm$2.04} & \textbf{80.80$\pm$4.94} & \textbf{3.26$\pm$2.15} \\
    \hline
    \hline
    \end{tabular}}
    \label{tab4:cardiac_ablation}
\end{table*}
\subsubsection{Comparison with Existing Methods}
We employed almost the same set of methods as in Section~\ref{subsubsec:vel_sota} for comparison in the experiment. 
As the the SC-GAN was specifically designed for vessel segmentation, we replace it with CycleGAN~\cite{zhu2017unpaired} for comparison here. CycleGAN is a commonly used image-level alignment UDA method that utilizes cycle-consistent generative adversarial networks to achieve cross-modality synthesis. We employed Dice score and Average Symmetric Surface Distance (ASSD) for quantitative evaluation.

Quantitative evaluation results of the compared methods are shown in Table~\ref{tab3:cardiac_sota}. It can be observed that Baseline (source) obtained a very low average Dice of $10.12\%$ for the LV and Myo segmentation, showing the large domain shift between retinal fundus images and CMR. Baseline (target) obtained an average Dice of $66.21\%$, which shows that only using the small number of annotated images in the target domain will limited the model's performance. For UDA methods, SIFA~\cite{chen2019synergistic} performed better than ADDA~\cite{tzeng2017adversarial} and CycleGAN~\cite{zhu2017unpaired}, but its average Dice was only $59.97\%$, showing that classical UDA methods cannot be directly applied to cross-anatomy domain adaptation. 
For SDA methods, Fine-tuning (last)~\cite{Tajbakhsh2016fine} had the worst performance with an average Dice of $35.75\%$ while fine-tuning (all) achieved an average Dice score of $78.63\%$. This is very impressive and it shows that just fine-tuning is effective to achieve a better performance than learning directly from the small set of target images. However, its performance is lower than our CS-CADA that improved the average Dice to $80.80\%$. For the values for Joint Training, X-shape~\cite{valindria2018multi} and DSBN~\cite{chang2019domain} were $66.32\%$, $75.41\%$ and $77.80\%$, respectively. The results shows the effectiveness of DSBN to deal with the domain gap. The
SSL methods are generally better than the UDA and SDA methods, and the average Dice for SE-MT~\cite{french2018selfensembling}, UA-MT~\cite{yu2019uncertainty} and CPS~\cite{chen2021semi} were $75.65\%$, $76.76\%$ and $74.96\%$, respectively. The excising SSDA method Dual-T~\cite{li2020dual} obtained an average Dice of $76.22\%$, which was similar to UA-MT and outperformed the other existing methods. In comparison, our proposed CS-CADA achieved higher performance than most above state-of-the-art methods, and it obtained an average Dice of $80.80\%$ by combining DSBN and our proposed cross-domain contrastive learning in a semi-supervised framework. The ASSD values obtained by our CS-CADA for the LV and Myo were $3.15mm$ and $3.37mm$, respectively, which were also superior in ASSD values of the other methods.

Fig.~\ref{fig6:cardiac_seg} shows a visual comparison between different methods for the LV and Myo segmentation, where we selected SIFA~\cite{chen2019synergistic}, DSBN~\cite{valindria2018multi} and UA-MT~\cite{yu2019uncertainty} that are the best UDA, SDA and SSL methods according to Table~\ref{tab3:cardiac_sota}, respectively. It indicates that Baseline (target) achieved very poor results, with over-segmentation of LV and some missing part of Myo. SIFA captured the shape of LV and Myo through style transferring, but the result does not match the semantic boundary well and the Myo is noticeably over-segmented. In contrast, DSBN and UA-MT can better predict the overall structure than these two methods, but they gain a lot of false positive predictions. %The reason may be that the ASBN just ease the domain shift while UA-MT designed for annotation problem.
Our CS-CADA achieved much better results than the others, without false positives in the background, and the segmentation boundary is very close to that of the ground truth. 
%In comparison, SS-CADA and CS-CADA achieve better performance in visual predictions compared with other methods. Although SS-CADA also has mis-predicted parts, its circumstance is better than ASBN and UA-MT. The visual comparison between SS-CADA and CS-CADA demonstrates that the contrastive learning will improve the network's ability of capturing domain-invariant features.
\begin{figure}[t]
    \centerline{\includegraphics[width=3in]{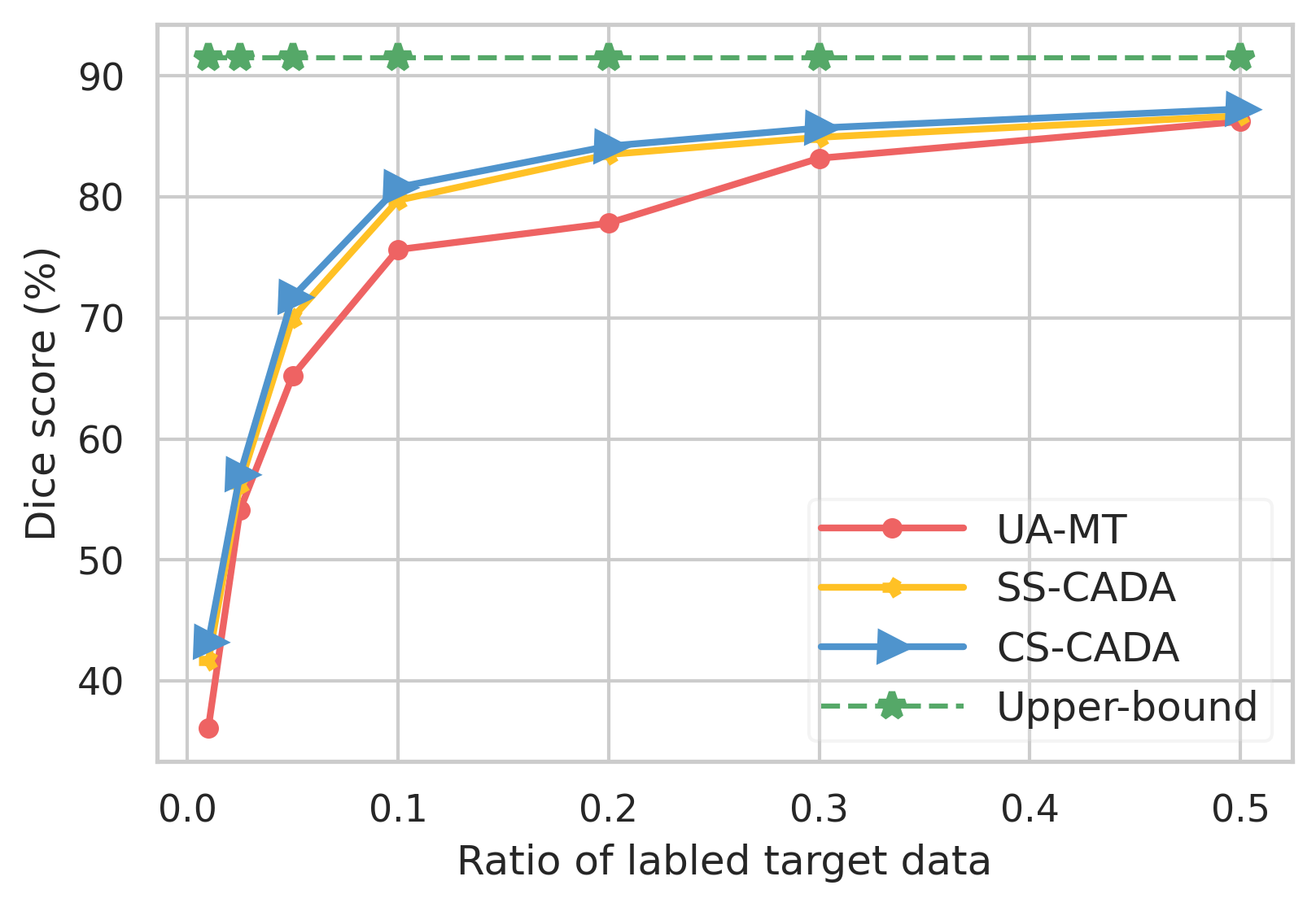}}
    \caption{Average Dice of LV and Myo achieved by different methods under different ratios of labeled images in the target domain. The green dotted line is the upper bound on fully supervised learning.}
    \label{fig7:cmr_ablations}
\end{figure}
\subsubsection{Ablations Study}
In parallel with Section~\ref{subsubsec:vel_abla}, we conducted the ablation study by comparing our CS-CADA with SE-MT that only leans from $T_L$ and $T_U$,  DSBN that only leans from $S_L$ and $T_L$, and  SS-CADA~\cite{zhang2021sscada} that combines SE-MT and DSBN but does not use $L_{ct}$. The evaluation results of these methods are shown in Table~\ref{tab4:cardiac_ablation}. Compared with Baseline (target), SE-MT and DSBN improved the average Dice from 66.21\% to 75.65\% and 77.80\%, respectively, which validates the effectiveness of them to leverage $T_U$ and $S_L$, respectively. SS-CADA combining them together further improved the score to 79.25\%. Finally, the better performance of CS-CADA than SS-CADA shows the effectiveness of our contrastive learning strategy. We also conducted statistical significance evaluation based on paired t-test between SS-CADA and CS-CADA. The p-value for the Dice score was 0.043 $<$ 0.05, showing the significant improvement from SS-CADA to CS-CADA. Note that with only 10\% of training images in the target domain being annotated, our CS-CADA improved the average Dice score by 22\% (from 66.21\% to 80.80\%).

We also compared our CS-CADA with UA-MT and SS-CADA under different ratios of annotated images in the target domain, and their average Dice scores are shown in 
Fig.~\ref{fig7:cmr_ablations}. It can be observed that CS-CADA consistently outperformed UA-MT and SS-CADA. The gap between UA-MT and CS-CADA is larger when the annotation ratio is smaller, indicating that the effectiveness of our method leveraging additional dataset with similar structures in a different domain improves the segmentation performance when annotations in the target domain are limited. 
\section{Discussion and Conclusion}
\label{sec5:discussion}
Reducing the annotation cost while maintaining the model's robust performance is important in medical image segmentation tasks because of the time-consuming and labor-intensive annotation process. Considering the availability of existing annotated datasets for a similar anatomical structure, our cross-anatomy domain adaptation method can improve the segmentation performance by transferring knowledge from such available datasets to the target segmentation task where only a small set of annotations are available for the target, which outperformed fine-tuning for transfer learning. Our method is also superior to existing semi-supervised methods that do not consider images from other domains, and better than existing domain adaptation methods that focus on the same anatomical structure in similar or different modalities. Note that in standard transfer learning setting, the pre-trained model is fine-tuned with the target dataset, without using the source dataset any more. If the source and target dataset have a large gap, the fine-tuning is less effective. In the DA setting of this work, the shared shape features can improve the performance on the target dataset effectively. Compared with pre-training with ImageNet, using datasets with similar shapes is more data and annotation-efficient.

Differently from classical domain adaptation considering only cross-modality domain shift for segmenting the same set of structures~\cite{chen2019synergistic,chen2021adaptive}, we deal with a more difficult scenario where an existing dataset with similar anatomical structures is used to assist model training in the target domain. Here, the “similar anatomical structure” requirement means there is a shape similarity between the two domains, i.e., the shapes have similar topologies but may be different in scales. For example, vessels in different organs are similar in terms of vascular shapes, but they can have different diameters and orientations. In the second scenario of our work, the LV and Myo in cardiac MRI and optic cup and disk in fundus images are obviously two different domains, but both of them have circular structures in different scales. %Because of the cooperating of the shared convolutional layer, DSBN and cross-domain contrastive learning, CS-CADA will comprehensively capture domain-invariant information while smoothly remain and map domain-specific distribution. CS-CADA allows larger style difference in the two domains compared with traditional DA methods, e.g., in our second task, retinal image as source domain and cardiac MRI as the target.

Differently from disentanglement methods~\cite{lee2018diverse,benaim2019domain}, the proposed CS-CADA does not need to specifically design modules to extract domain-invariant content and domain-specific style representations, respectively. Our CS-CADA captures anatomical representations through the shared convolutional layers and normalizes each style distribution to a common distribution space by DSBN. The whole procedure is conducted on a unified network. In contrast, disentanglement methods typically need to disentangle out the content and style representations through different networks and additionally introduce generative adversarial networks to discriminate them. These procedures are more complex and difficult to train compared with CS-CADA.

The term “across similar anatomical structures” is introduced for the aim of improving the model’s performance on a target domain with the help of some easily available datasets with similar structures. Indeed, requiring “similar structures” is more relaxed than the requirement of the exact set of objects in different domains (like CT and MRI) in typical DA setting. Though our method can also be applied to the same set of objects in different domains, this work makes it possible to adapt a model to a similar structure, rather than the same structure, which increases the chance of leveraging broader source domains. Compared with cross-domain contrastive learning, KL loss can also be used to encourage the distributions of $g_i^\mathcal{S}$ and $g_j^\mathcal{T}$ to be similar, but at the same time, we aim to encourage ($g_i^\mathcal{S}$, $g_i^\mathcal{T}$) and ($g_i^\mathcal{S}$, $g_j^\mathcal{S}$) to be divergent so that the BN layers can distinguish the different styles of two domains. Hence, using contrastive learning is more suitable for this purpose.

One limitation of this work is that it requires some annotated samples in the target domain for model training. As shown in Fig. 4 and Fig. 6, our proposed CS-CADA performed well when the annotation ratio changes from $30\%$ to $50\%$, but in extreme situations (e.g., the annotation ratio is less than $10\%$), its performance drops dramatically. It may be a potential way to combine our method with a small-sample learning strategy (e.g., few/one-shot learning, test-time adaptation, etc.) to improve model adaptation ability in such situations. In addition, this work only deals with 2D images, and it is of interest to deal with 3D structures in the future.

In conclusion, we propose a Contrastive Semi-supervised learning for Cross Anatomy Domain Adaptation (CS-CADA) in medical image segmentation with knowledge transferred from an existing dataset to a target segmentation task, where they are different organs with similar anatomical structures. We use Domain-Specific Batch Normalization (DSBN) and shared convolutional kernels to jointly learn from the source and target domains, and propose cross-domain contrastive learning to learn domain-invariant features, which is combined with a self-ensembling mean teacher framework to further leverage unannotated images in the target domain.
%The proposed CS-CADA is compact and flexible without using adversarial learning with additional network designs and the complex training process.
% We assumed that there is a modest number of dataset with only limited annotations are available in the medical imaging scenario. In our setting, classical annotation-efficient methods could not suitable for training a satisfactory network because two obstacles (limited annotation and domain adaptation) were not be considered previously. 
%Our proposed CS-CADA can mainly summarised as three parts: 1) self-ensembling  mean-teacher as backbone; 2) anatomy-specific batch normalization for specific feature distribution; and 3) contrastive learning for capturing consistent structure content across domain. Three components overcome the challenging cross-anatomy domain shift and facilitate the annotation efficiency. 
Experimental results show that our method can effectively achieve accurate segmentation of coronary artery from XAs and left ventricle and myocardium from MRI with limited annotations, which has a potential to reduce the annotation cost.
In the future, it is of interest to apply our method to other segmentation tasks, and improve the robust performance under the extremely limited annotations.

% \section*{References and Footnotes}

% \section*{References}
% \label{}
\bibliographystyle{ieeetr} 
\bibliography{IEEEabrv,refs}

\end{document}